\documentclass[sigconf, nonacm]{acmart}

\makeatletter
\renewcommand{\@authorfont}{\small}
\renewcommand{\@affiliationfont}{\footnotesize}
\makeatother
\usepackage{booktabs} 

\setcopyright{rightsretained}
\usepackage{caption}
\usepackage{subcaption}
\usepackage{microtype}
\usepackage[shortcuts]{extdash}

\begin{document}
\title{ODG-NoMaD: Overhead-Camera Direction-Guided NoMaD}

\author{Blossom Treesa Bastian, Keerthi S. Shetty, Manish Kolachalam, Rani Malhotra,Ashish Dutta    }
   \thanks{Corresponding to blossomtreesa.b@infosys.com}
\thanks{Blossom Treesa Bastian, Keerthi S. Shetty, Manish  Kolachalam, Rani Malhotra are with Applied Research Center for Autonomous Machines, Infosys Center for Emerging Technologies, Banglore, India.}%
\thanks{Ashish Dutta is with Dept. of Mechanical Engineering, IIT Kanpur, Kanpur, India.}%

\renewcommand{\shortauthors}{}

\begin{abstract}

Learned diffusion policies such as NoMaD unify goal-conditioned navigation and
exploration in a single model, but in an unseen environment---with no goal image
or topological map---they can only explore undirectedly, wandering without global
awareness. We present ODG-NoMaD, which combines the global guidance of classical
planning with the reactivity and generalization of NoMaD's exploration mode,
without retraining the policy. An overhead depth camera is used once on deployment
to build an occupancy map and plan a global path, which is segmented to yield a
desired heading. A per-frame traversability map from the robot's onboard depth
then refines this heading into a collision-free direction, and the trajectory
samples of NoMaD's diffusion policy are steered toward it by injecting the gradient
of a direction cost into the denoising process. Experiments in simulated office environments show that ODG-NoMaD reduces the mean
residual distance to the target by roughly $7\times$ relative to unguided
exploration and outperforms the point-goal cost guidance of NaviDiffusor. It further avoids collisions with new obstacles introduced after the
overhead map is built, deviating around them while still following the global path
to the goal.

\end{abstract}

%
%

\keywords{Visual navigation, Diffusion policies, Guided exploration, Overhead-camera mapping}

\maketitle

\section{Introduction}
\label{sec:intro}

Autonomous navigation in unknown or partially known environments is a core capability for mobile robots, underpinning applications as varied as disaster-zone inspection, operation in hazardous settings, search-and-rescue, and last-mile delivery. The challenge is sharpest when a robot must explore and move safely relying solely on its own onboard observations. Two broad families of methods address this problem. Classical autonomous navigation pipelines decompose the task into sequential stages of perception, mapping, localization, and planning, generating collision-free paths by optimizing hand-crafted cost functions over an explicitly reconstructed map \cite{marder2010office,hornung2013octomap,hess2016cartographer,mur2015orb}. While this modular design provides global situational awareness and interpretable decision-making, it is prone to error accumulation across stages and relies heavily on maintaining an accurate metric map.  In contrast to classical pipelines, learning-based navigation methods directly map raw sensory observations, particularly visual inputs, to control actions, learning navigation priors from large and diverse datasets \cite{shah2023gnm,shah2023vint,sridhar2024nomad}. By eliminating the need for explicit map construction, these approaches offer greater flexibility and can generalize across different robotic platforms. However, their performance often degrades in out-of-distribution environments, and the absence of an explicit environmental representation limits their ability to exploit the global spatial structure that classical planning methods naturally leverage.

NoMaD \cite{sridhar2024nomad} exemplifies the strengths of the learning-based paradigm, serving as a versatile vision-based navigation foundation model. Through goal masking, a single diffusion policy performs both goal-conditioned navigation and open-ended exploration: when a goal image is supplied it drives toward it, and when no goal image is given it explores by sampling from a diverse, multimodal distribution of collision-aware action candidates produced by a trained action-diffusion policy.  
However, NoMaD's exploration mode is fundamentally \textit{undirected}: while it yields locally diverse and feasible trajectories, these generated trajectories carry no global situational awareness, and the policy's decisions are egocentric and reactive, constrained by the limited field of view and occlusions of the onboard camera. As a result, exploration in an unfamiliar space is often inefficient—revisiting already-covered regions, wandering without clear progress, discovering new frontiers slowly, and becoming trapped in local pockets of the environment~\cite{dreamNav}. This is precisely the global awareness that classical map-based planners provide~\cite{yamauchi1997, gonzalezbanos2002}—and that purely reactive learned exploration discards~\cite{chaplot2020}.

\begin{figure}[t]
  \centering
  \includegraphics[width=\columnwidth]{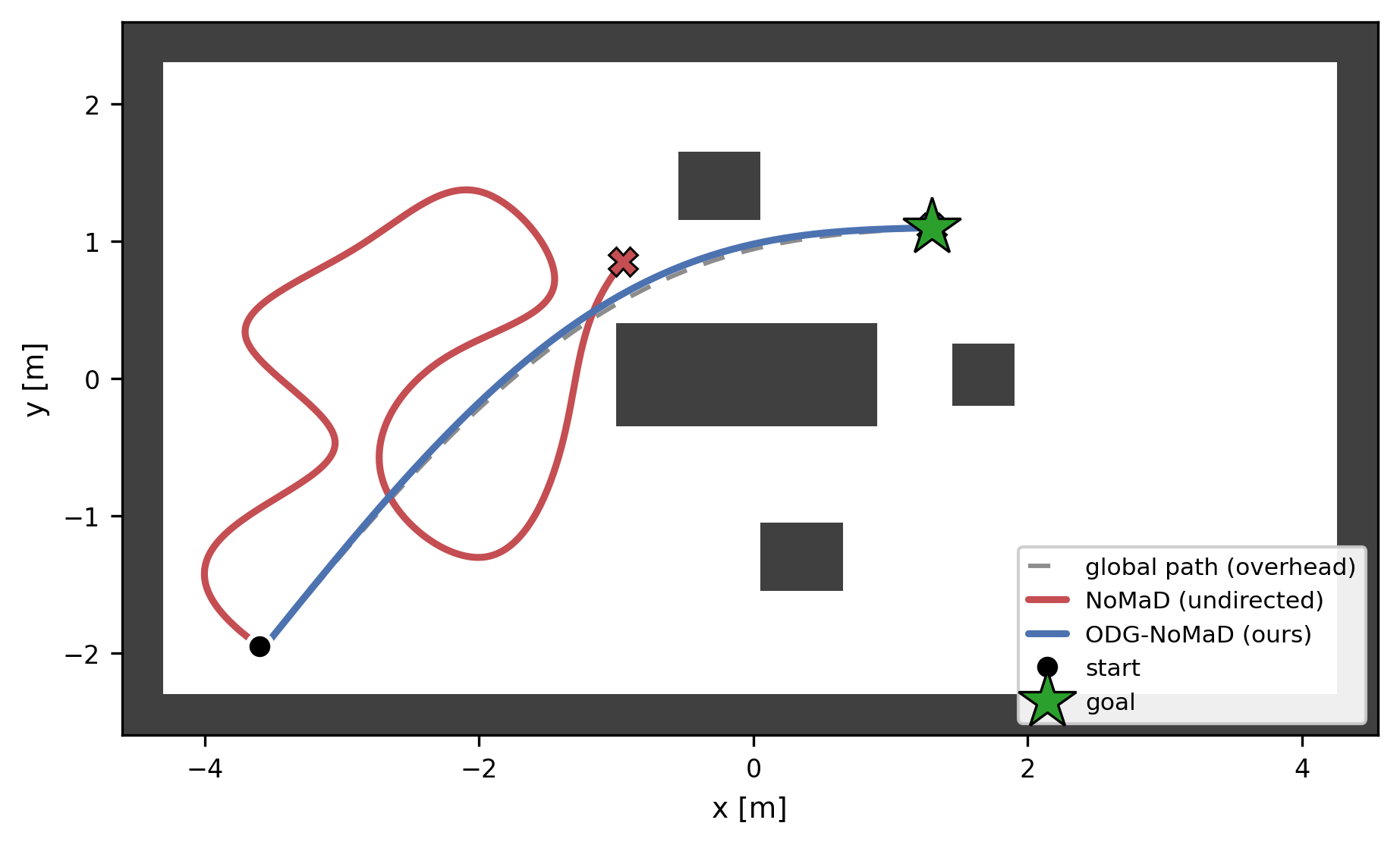}
  \caption{Exploration in a previously unseen environment, shown on the
  occupancy grid built from the overhead camera (black: obstacles and walls,
  white: free space). Both runs begin at the same start ($\bullet$) and target
  the same goal ($\star$). In its default exploration mode, NoMaD (red) is
  \emph{undirected}: with no global situational awareness it wanders, revisits
  already-covered regions, and halts far from the target ($\times$). ODG-NoMaD
  (blue) biases the same goal-masked policy along a global path planned on the
  overhead-camera map (grey dashed), steering exploration to the goal along a
  markedly shorter route.}
  \label{fig:teaser}
\end{figure}

At this juncture, one promising direction is to combine the global guidance of classical planning with the reactivity and generalization of a learned policy, without sacrificing either. This can be achieved by introducing external, globally-aware signals into the inference stage of a navigation diffusion policy, leaving the trained network untouched. NaviDiffusor~\cite{navidiffusor}, for instance, guides the denoising process of an action-diffusion policy with differentiable scene- and task-level cost gradients to improve goal-reaching and collision avoidance in a plug-and-play manner, without retraining.
We adopt this inference-time guidance philosophy, but obtain the guiding direction \textit{autonomously} and \textit{globally}. We propose \textbf{Overhead-Camera Direction-Guided NoMaD (ODG-NoMaD)}, an exploration method for unseen environments in which a global planner is run once on an initial map built from an overhead camera. The resulting global path, leading from the robot's current position to the target, defines a direction that is used as a directional prior to re-weight the candidate trajectories sampled by NoMaD's goal-masked diffusion policy---biasing exploration toward that heading while preserving the policy's reactive, collision-aware behavior. This changes the exploration behavior substantially: as Fig.~\ref{fig:teaser} shows, from an identical start and goal, unguided NoMaD wanders and halts far short of the target, whereas ODG-NoMaD follows the global path and reaches the goal along a markedly shorter route.
We further compare this strategy with the goal-guided approach of NaviDiffusor~\cite{navidiffusor}, whose differentiable goal cost steers diffusion sampling to a specified point. We find that direction guidance is superior at reaching the final point of the global path. We attribute this to the nature of exploration: a heading prior gently biases NoMaD while preserving its exploratory multimodality, whereas a precise point target collapses the action distribution toward a fixed location.
Finally, to improve the collision avoidance of the learned policy, we compute a traversability map from the depth image and use it to select a heading that steers the robot away from obstacles that are out-of-distribution for the learned policy, while remaining aligned with the desired heading from the global path. 
The main contributions are:
\begin{itemize}
\item \textbf{ODG-NoMaD}, an exploration framework that guides NoMaD's goal-masked diffusion policy at inference time with a directional prior derived from a global path over an overhead-camera view of the workspace, steering exploration towards the target without a human in the loop and without retraining the policy.
\item A traversability-map-based collision-avoidance scheme that steers the robot away from obstacles that are out-of-distribu\-tion for the learned policy while keeping it aligned with the global-path heading.
\item A head-to-head comparison against NaviDiffusor's goal-cost guidance, showing our directional prior is the more reliable way to inject the global path.

\item Robustness to new objects or obstacles introduced into the environment
\emph{after} the overhead map is created: these are avoided while ODG-NoMaD
follows the global path to the goal without collision.
\end{itemize}
\vspace{-0.75\baselineskip}
\section{Related Work}
\label{sec:related}
 
We organize the related literature along the two threads that ODG-NoMaD brings
together. Section~\ref{sec:rw_overhead} reviews navigation systems that exploit
an external, overhead view of the workspace---typically ceiling- or
infrastructure-mounted cameras---for localization, mapping, and planning, which
motivates our use of a top-down camera for global situational awareness.
Section~\ref{sec:rw_nomad} then reviews methods that guide or modify the
trajectories produced by NoMaD and related diffusion navigation policies,
predominantly at inference time, which motivates how we inject that awareness
into the policy. 
\subsection{Infrastructure and Overhead-Camera-Based Navigation}
\label{sec:rw_overhead}
 
External, infrastructure-mounted sensing has long been used to localize and
guide mobile robots whose onboard perception is limited. Overhead and
ceiling-mounted cameras are especially common: multi-camera ceiling rigs track
and localize one or more robots simultaneously in a shared world frame
\cite{7370454}, and networks of fixed indoor surveillance cameras
estimate robot and object positions through image-to-ground homographies
\cite{s16020195}. For smaller robots that cannot rely on GPS modules, \cite{article1} employed an overhead camera combined with robot-mounted LEDs for simultaneous tracking and identification. Beyond pure localization,
distributed networks of smart overhead cameras have been used for decentralized
map building and hierarchical path planning for autonomous ground vehicles
\cite{iet-cvi.2019.0949}, whereas \cite{REKLEITIS2006921} treats a network of cameras as a localization source, using robot-mounted fiducial markers and an Extended Kalman filter to jointly calibrate the cameras as a mobile robot explores the network to plan and navigate.
 
The recurring motivation is that an overhead viewpoint provides global, top-down
situational awareness---free space, obstacles, and the robot's pose in a world
frame---that a first-person onboard camera cannot recover because of its limited
field of view and frequent occlusions. However, these systems almost universally
route the overhead information into \emph{classical} localization and planning
stacks: the top-down view drives metric mapping and geometric planners rather
than a learned reactive policy. Our approach preserves the global overhead
advantage but couples it, for the first time, to a learned diffusion navigation
policy.

\subsection{Steering and Modifying NoMaD's Diffusion Trajectories}
\label{sec:rw_nomad}
 
Diffusion models have become a powerful mechanism for generating robot
trajectories \cite{diffusionpolicy}. In navigation, ViNT \cite{shah2023vint} established
a Transformer foundation model for image-goal navigation, and NoMaD
\cite{sridhar2024nomad} unified goal-conditioned navigation and undirected exploration
within a single goal-masked diffusion policy that operates directly from onboard
RGB. A growing body of work builds on NoMaD by \emph{shaping or steering} its
generated trajectories rather than retraining the policy from scratch.
 
The most closely related is NaviDiffusor \cite{navidiffusor}, which augments each
denoising step with the gradient of differentiable scene- and task-level cost
functions, biasing the sampled path toward collision-free, goal-reaching
trajectories in a plug-and-play fashion, and which explicitly adopts NoMaD as a
baseline. This inference-time, cost- and constraint-guided sampling paradigm is
shared by a broader family of trajectory generators that inject barrier or
constraint gradients into the reverse diffusion process
\cite{codig, cgd, mpd}. Complementary efforts extend the diffusion-navigation
line in other directions: local diffusion planners for efficient collision
avoidance \cite{ldp}, diffusion-based autonomous exploration planners that reason
over belief maps \cite{dare}, sim-to-real navigation diffusion with
privileged-information guidance \cite{navdp}, and image-space diffusion adapted
for task-conditioned navigation \cite{ventura}. Closest in spirit to our
directional guidance, Nakaoka et al.\ \cite{nakaoka} condition a diffusion
navigation policy on directional cues---human pointing gestures and arrow
signage---interpreted by a vision-language model.
 
Across all of these works, the signal that shapes the trajectory is supplied
either by a human in the loop \cite{nakaoka} or by costs defined over the robot's
own \emph{egocentric} observations \cite{navidiffusor, codig, cgd}. None derives
that guidance from a global, overhead view of the environment. Our work sits at the intersection of these two threads. From the first
(Sec.~\ref{sec:rw_overhead}) we inherit the overhead camera's global, top-down
situational awareness; from the second (Sec.~\ref{sec:rw_nomad}) we inherit the
principle of steering a NoMaD diffusion policy at inference time, without
retraining. 

\section{Overhead-Camera Direction-Guided NoMaD}
\label{sec:method}

\begin{figure*}[t]
  \centering
  \includegraphics[width=\textwidth]{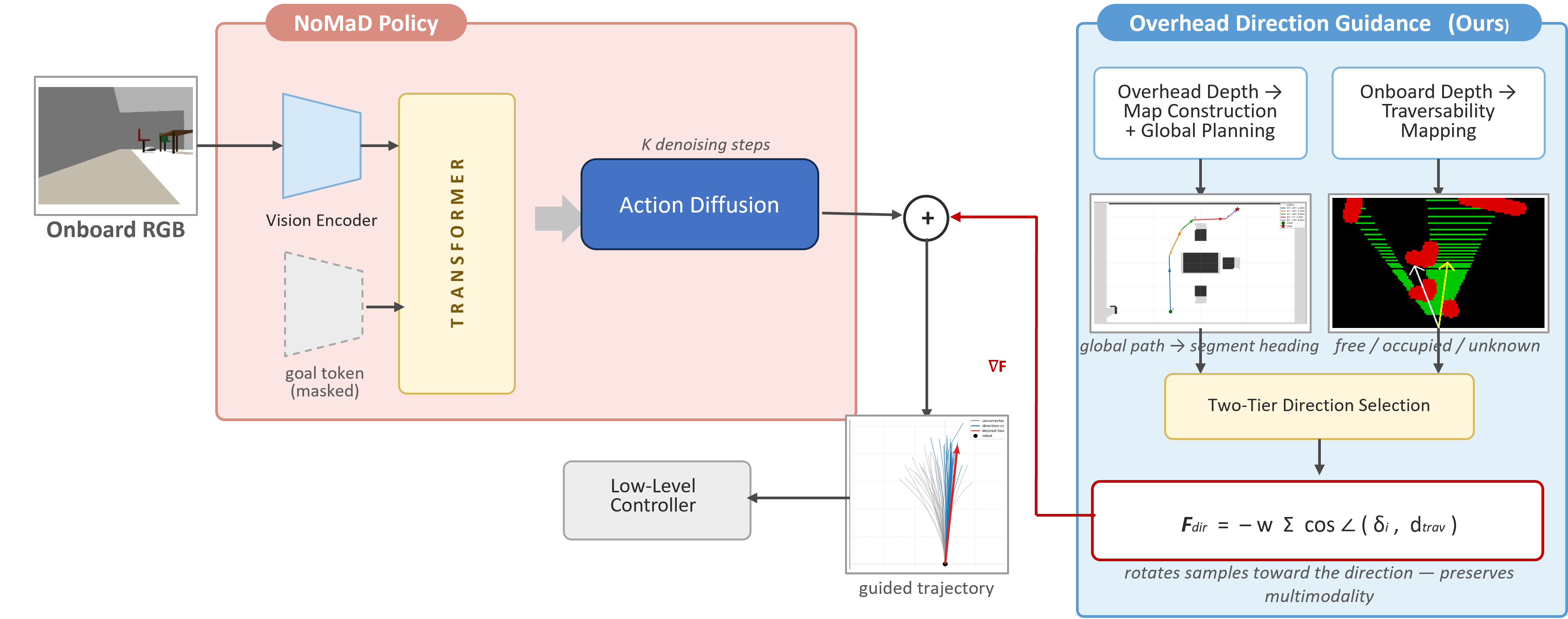}
  \caption{Overview of ODG-NoMaD. The pretrained NoMaD policy (left) encodes the
  robot's onboard RGB observation and, with its goal token masked, samples a
  multimodal set of trajectory candidates through the action diffusion model
  $\Phi_\theta$ over $K$ denoising steps. Our guidance module (right) supplies the
  missing global context: an overhead depth camera is used once on deployment to
  build an occupancy map and plan a global path, which is segmented to yield a
  desired heading, while the robot's onboard depth stream is used per frame to
  build a BEV traversability map. A two-tier selector reconciles the two into a
  collision-free direction $\hat{\mathbf{d}}_{\mathrm{trav}}$, which defines the
  direction cost $F_{\mathrm{dir}}$ of Eq.~\eqref{eq:dircost}. Its gradient
  $\nabla F_{\mathrm{dir}}$ is injected into the denoising process
  (Eq.~\eqref{eq:guided_update}), rotating the samples toward the desired
  direction while preserving their spread.}
  \label{fig:architecture}
\end{figure*}

In this section we detail the ODG-NoMaD pipeline. We first describe how a metric
map of the workspace is built from the depth image provided by the overhead
camera and how a global planner computes a path from point $A$ to point $B$ in
world space on this map (Sec.~\ref{sec:mapping}). We then explain how this global
path is segmented to derive the desired direction of navigation
(Sec.~\ref{sec:segmentation}). Finally, we describe how this direction, combined
with a collision-avoidance term, guides---and thereby modifies---the trajectories
generated by NoMaD's action-diffusion model(Sec.~\ref{sec:guidance}). Fig.~\ref{fig:architecture} gives an overview of ODG-NoMaD.

\subsection{Map Construction and Global Path Planning}
\label{sec:mapping}

The overhead depth camera is mounted looking down on the workspace, and a
one-time extrinsic calibration provides the rotation $R_{\text{floor}}\in SO(3)$
that aligns the camera optical frame with the floor plane, together with the
camera height $H_{\text{cam}}$ above the floor. On deployment, a single depth frame is converted into a top-down occupancy grid
as follows. Valid pixels ($0.3\,\text{m} < d < 8\,\text{m}$) are back-projected
into a metric point cloud $\mathbf{p}_c = (x_c,\,y_c,\,z_c)$ using the pinhole
model~\cite{hartley2004} and the camera intrinsics, rotated into the world frame
as $\mathbf{p}_w = (x_w,\,y_w,\,z_w) = R_{\text{floor}}\,\mathbf{p}_c$, and
assigned a height above the floor $h = H_{\text{cam}} - z_w$.

\begin{figure}[t]
  \centering
  \begin{subfigure}[b]{0.49\columnwidth}
    \centering
    \includegraphics[width=\linewidth]{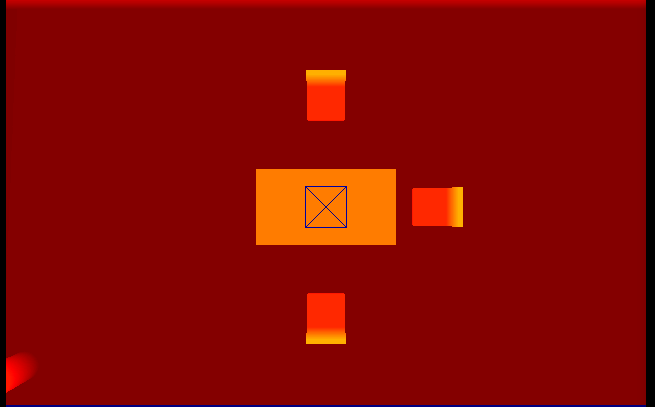}
    \caption{Overhead depth image}
    \label{fig:mapping_depth}
  \end{subfigure}\hfill%
  \begin{subfigure}[b]{0.49\columnwidth}
    \centering
    \includegraphics[width=\linewidth]{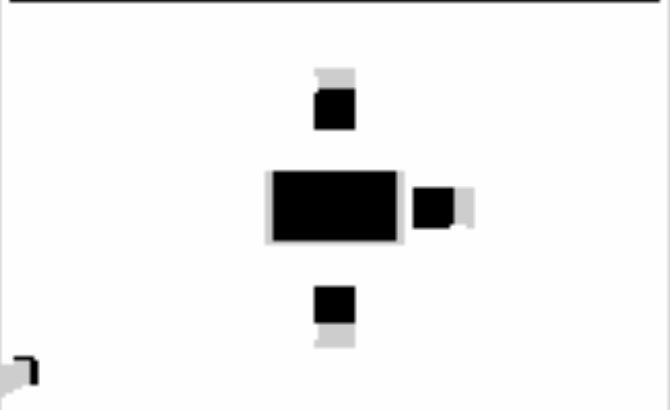}
    \caption{Occupancy grid map}
    \label{fig:mapping_grid}
  \end{subfigure}
  \caption{Map construction from the overhead camera.
  (a)~The colourised depth image observed from the ceiling-mounted camera.
  (b)~The occupancy grid obtained from the depth image.
  Black denotes occupied cells, white free cells, and grey unknown cells.}
  \label{fig:mapping}
\end{figure}

Points are then classified purely by height into a top-down grid, in the spirit
of elevation-based $2.5$D occupancy mapping~\cite{elfes1989, thrun2005}: a point
is \emph{free} within a thin floor band, \emph{occupied} between the floor band
and a ceiling cut-off (which rejects the ceiling and the sensor itself), and
\emph{unknown} elsewhere, using thresholds
$(h_{\text{lo}},\,h_{\text{hi}},\,h_{\text{ceil}}) =
(-0.05,\,0.08,\,2.5)\,\text{m}$. Labelled points are rasterised at
$r = 0.05\,\text{m}$ resolution over the workspace footprint observed by the
overhead camera, and the occupied cells are despeckled with a morphological
closing followed by an opening ($3\times3$ elliptical
element)~\cite{gonzalez2018}, yielding the occupancy grid $M$.
Fig.~\ref{fig:mapping} shows an example depth image and the resulting map.

On this grid, a grid-based global planner~\cite{nav2} computes a collision-free
path $\mathcal{P} = \{\mathbf{p}_0,\dots,\mathbf{p}_N\}$ in world coordinates from
the robot's current position $A=\mathbf{p}_0$ to the target $B=\mathbf{p}_N$,
which serves as the global reference for the direction extraction of
Sec.~\ref{sec:segmentation}.
 
\subsection{Path Segmentation and Direction Extraction}
\label{sec:segmentation}
The global path $\mathcal{P} = \{\mathbf{p}_0,\dots,\mathbf{p}_N\}$ returned by the
planner is a dense polyline whose per-waypoint headings are noisy and therefore
unsuitable as a direct guidance signal. We simplify $\mathcal{P}$ into a small set
of near-straight segments of approximately constant heading using the
Ramer--Douglas--Peucker (RDP) algorithm \cite{ramer1972, douglaspeucker1973}.
Given a sub-polyline delimited by endpoints $\mathbf{p}_0$ and $\mathbf{p}_N$, RDP
locates the intermediate waypoint of maximum perpendicular distance to the chord
$\overline{\mathbf{p}_0\mathbf{p}_N}$,
\begin{equation}
d_\perp(\mathbf{p}_i;\mathbf{p}_0,\mathbf{p}_N) =
\frac{\big\lvert (\mathbf{p}_b - \mathbf{p}_0)\times(\mathbf{p}_i - \mathbf{p}_0)\big\rvert}
     {\lVert \mathbf{p}_N - \mathbf{p}_0 \rVert},
\label{eq:rdp_dist}
\end{equation}
retains it and recurses on both halves whenever $d_\perp > \epsilon$, and
otherwise discards all intermediate waypoints. The tolerance $\epsilon$ (in
metres) trades fidelity against the number of segments. The retained waypoints
form an ordered set of $M+1$ breakpoints $0 = b_0 < b_1 < \dots < b_{M-1} = N$, and
consecutive breakpoints delimit the segments
$\mathcal{S}_k = (\mathbf{p}_{b_k}, \mathbf{p}_{b_{k+1}})$, $k = 0,\dots,M-1$.
 
\begin{figure}[t]
  \centering
    \includegraphics[width=\linewidth]{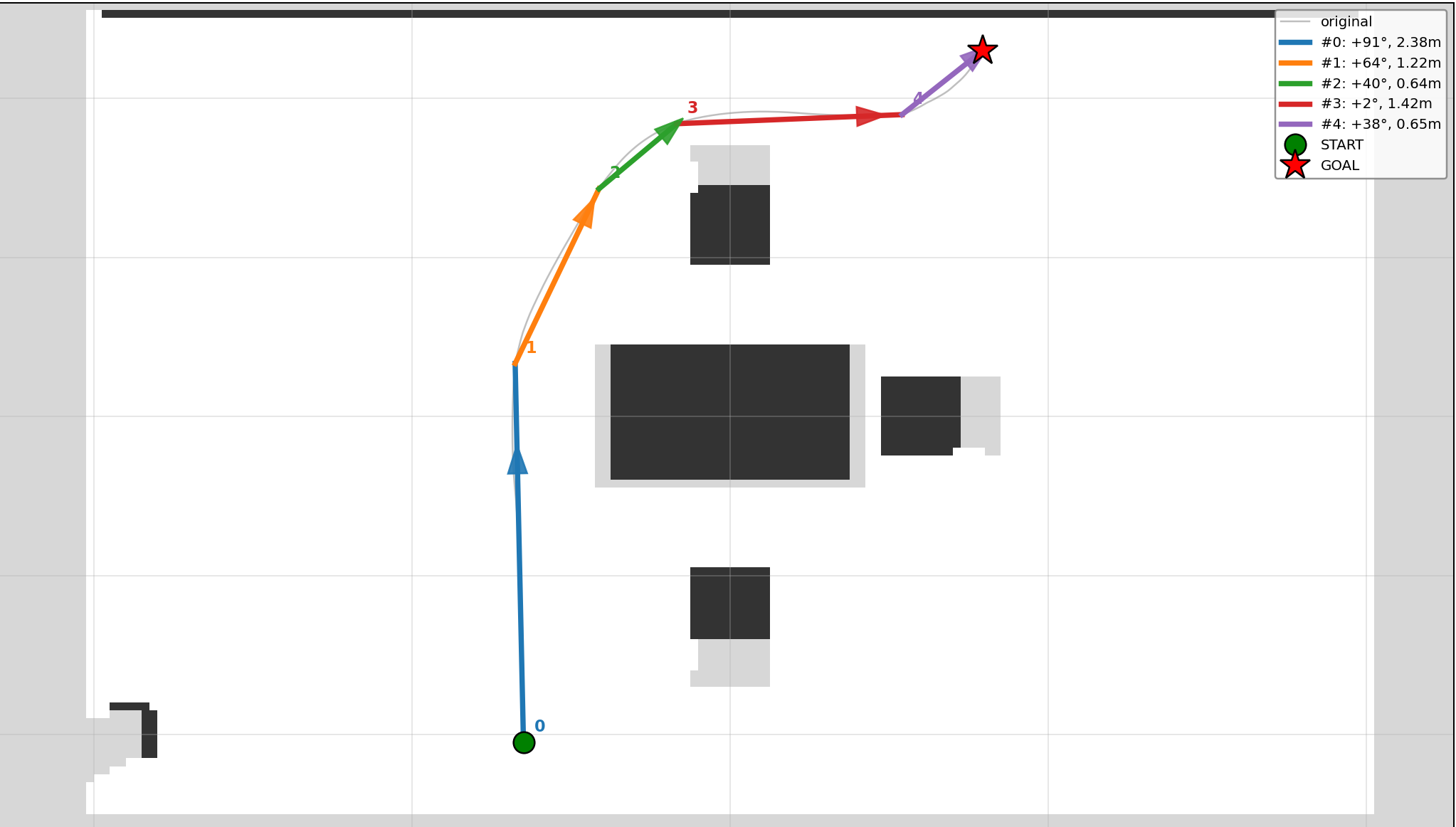}
  \caption{The dense polyline $\mathcal{P}$ and the segments produced by RDP simplification.}
  \label{fig:segmentation}
\end{figure}

Each segment is summarised by a single heading and direction. Writing
$\Delta x_k = x_{b_{k+1}} - x_{b_k}$ and $\Delta y_k = y_{b_{k+1}} - y_{b_k}$, the
segment heading and unit direction vector are:
\begin{equation}
\begin{aligned}
\theta_k &= \operatorname{atan2}(\Delta y_k,\,\Delta x_k), \qquad
\hat{\mathbf{d}}_k = (\cos\theta_k,\,\sin\theta_k)
\end{aligned}
\label{eq:segdir}
\end{equation}
The heading of the segment containing the robot's current position defines the
desired direction of travel $\theta^\star$ that is passed to the guidance stage
(Sec.~\ref{sec:guidance}). Fig.~\ref{fig:segmentation} shows an example input path together with the resulting segments, each annotated with its unit direction vector $\hat{\mathbf{d}}_k$ and heading $\theta_k$.
 
\subsection{Direction- and Collision-Guided Trajectory Modification}
\label{sec:guidance}

Our local navigation policy builds on NoMaD~\cite{sridhar2024nomad}, a
unified goal-masked diffusion policy for visual navigation. NoMaD extends
the ViNT foundation model~\cite{shah2023vint} with a binary goal-masking
mechanism: a Transformer encoder fuses the current and recent RGB
observations with an optional goal image, and a learned mask determines
whether the goal token participates in the fused context. When the goal
token is masked, the policy operates in \emph{exploration mode}: it is
conditioned only on the observation history and produces
\emph{task-agnostic}, multimodal candidate trajectories that capture the
diverse short-horizon motions afforded by the scene. This
undirected, multimodal behavior is precisely what makes exploration mode
attractive as a base policy; however, it provides no mechanism to bias
the samples toward a preferred direction of travel, nor any explicit
notion of obstacles beyond what is implicit in the training data. We
therefore steer the exploration-mode samples at inference time, without
any retraining, by combining (i)~gradient-based \emph{direction guidance}
applied inside the diffusion sampling loop
(Sec.~\ref{sec:direction_cost}) and (ii)~an explicit
\emph{traversability-map-based} direction selector that enforces
collision avoidance (Sec.~\ref{sec:traversability}).
 
\paragraph{Action diffusion in NoMaD}
NoMaD's action head is a conditional denoising diffusion probabilistic
model (DDPM)~\cite{ho2020ddpm}, following the Diffusion Policy
formulation of visuomotor control~\cite{diffusionpolicy}. The
policy represents a trajectory as a sequence of normalized waypoint
displacements $\mathbf{a} \in [-1,1]^{T \times 2}$, where
$T$ is the prediction horizon (the number of future waypoints the policy outputs at each step). Starting from
Gaussian noise $\mathbf{a}_K \sim \mathcal{N}(\mathbf{0}, \mathbf{I})$,
the model iteratively denoises the sample over $K$ steps. Each reverse
step uses a noise-prediction network $\varepsilon_\theta$ conditioned on
the fused observation context $c_t$:
\begin{equation}
\mathbf{a}_{k-1} \;=\;
\alpha \Bigl( \mathbf{a}_k - \gamma\,
\varepsilon_\theta\!\left(c_t,\, \mathbf{a}_k,\, k\right) \Bigr)
\;+\; \sigma\, \mathbf{z}, \qquad
\mathbf{z} \sim \mathcal{N}(\mathbf{0}, \mathbf{I}),
\label{eq:ddpm_step}
\end{equation}
where $\alpha$, $\gamma$, and $\sigma$ follow the DDPM noise
schedule. Because sampling is iterative, the intermediate trajectory
$\mathbf{a}_k$ can be perturbed \emph{between} denoising steps.
Analogously to classifier guidance~\cite{dhariwal2021diffusion}, and
following cost-guided navigation diffusion~\cite{navidiffusor},
we replace the learned classifier gradient with the gradient of an
explicit, differentiable cost $F(\mathbf{a})$ and apply, during the final
(low-noise) denoising steps,
\begin{equation}
\mathbf{a}_k \;\leftarrow\;
\operatorname{clip}\!\bigl(\mathbf{a}_k - \eta\, \nabla_{\mathbf{a}_k}
F(\mathbf{a}_k),\, -1,\, 1\bigr),
\label{eq:guided_update}
\end{equation}
where $\eta$ is a guidance step size and the clipping keeps the sample
inside the action range seen during training. The guidance is
training-free: the diffusion model remains frozen, and all directional
and collision behavior is injected purely at inference time.
 
\subsubsection{Direction  Guidance of Sampled Trajectories}
\label{sec:direction_cost}
 
Exploration-mode samples are undirected by construction. To bias them
toward a desired direction of travel, we define a per-step \emph{alignment}
cost and use its gradient in the guided update of
Eq.~\eqref{eq:guided_update}. The desired direction
$\mathbf{d}^{\mathrm{rel}} \in \mathbb{R}^2$ is the direction, expressed in
the robot body frame ($x$~forward, $y$~left), from the robot's current
position to the active intermediate waypoint of the global path obtained in
Sec.~\ref{sec:segmentation}; its normalized form
$\hat{\mathbf{d}} = \mathbf{d}^{\mathrm{rel}} / \lVert \mathbf{d}^{\mathrm{rel}} \rVert$
is the unit direction used in the cost. The robot pose required for this
transformation is tracked incrementally: given the robot's known initial
position in world coordinates, its subsequent pose is obtained by relative
tracking (odometry) of its motion, and the world-frame target direction is
then rotated into the current body frame.
 
The network operates on normalized displacements; the cost, however,
must be evaluated on metric geometry. Let $g(\cdot)$ denote the
differentiable de-normalization that maps a normalized delta
$\mathbf{a}^{(j)}$ back to a metric displacement using the per-dimension
training statistics $(\boldsymbol{\delta}_{\min}, \boldsymbol{\delta}_{\max})$,
\begin{equation}
\boldsymbol{\delta}_j \;=\; g\!\left(\mathbf{a}^{(j)}\right)
\;=\; \tfrac{1}{2}\bigl(\mathbf{a}^{(j)} + \mathbf{1}\bigr)
\odot (\boldsymbol{\delta}_{\max} - \boldsymbol{\delta}_{\min})
+ \boldsymbol{\delta}_{\min}.
\label{eq:denorm}
\end{equation}
A cumulative sum then converts these displacements into waypoint
positions relative to the robot,
\begin{equation}
\mathbf{p}_i \;=\; \sum_{j=1}^{i} \boldsymbol{\delta}_j
\;=\; \sum_{j=1}^{i} g\!\left(\mathbf{a}^{(j)}\right),
\qquad i = 1,\dots,T,
\label{eq:cumsum}
\end{equation}
with $\mathbf{p}_0 = \mathbf{0}$ (the robot origin), so that the per-step
displacements are recovered as $\boldsymbol{\delta}_i = \mathbf{p}_i - \mathbf{p}_{i-1}$.
The direction cost measures the \emph{cosine} alignment of every step
with the desired unit direction $\hat{\mathbf{d}}$:
\begin{equation}
F_{\mathrm{dir}}(\mathbf{a}) \;=\;
-\, w \sum_{i=1}^{T}
\frac{\boldsymbol{\delta}_i^{\!\top} \hat{\mathbf{d}}}
     {\lVert \boldsymbol{\delta}_i \rVert + \epsilon},
\label{eq:dircost}
\end{equation}
with weight $w$ (we use $w = 0.05$) and a small $\epsilon$ for numerical
stability. Two properties of Eq.~\eqref{eq:dircost} are essential.

First, \emph{the cost rotates each step toward $\hat{\mathbf{d}}$ without
lengthening it.} Because the cosine normalizes out the step length
$\lVert\boldsymbol{\delta}_i\rVert$ and is bounded in $[-1,1]$, the gradient of
$F_{\mathrm{dir}}$ with respect to each displacement is, to first order,
\emph{perpendicular} to that displacement---it changes the step's direction but
not its magnitude. This is why we use a cosine rather than a raw projection
$-\sum_i \boldsymbol{\delta}_i^{\!\top}\hat{\mathbf{d}}$: the raw projection is
unbounded and is minimized simply by making steps longer, which we found
empirically inflates the trajectory norm and pushes samples outside the trained
action range.

Second, \emph{the whole trajectory is reshaped coherently, not just its
endpoint.} Since each waypoint position $\mathbf{p}_i$ is a cumulative sum of all
preceding displacements (Eq.~\eqref{eq:cumsum}), the gradient
$\nabla_{\mathbf{a}} F_{\mathrm{dir}}$ back-propagates to every element of the
normalized action $\mathbf{a}$, so a single guidance step adjusts the entire
predicted path at once.

\begin{figure}[t]
  \centering
  \includegraphics[width=0.9\columnwidth]{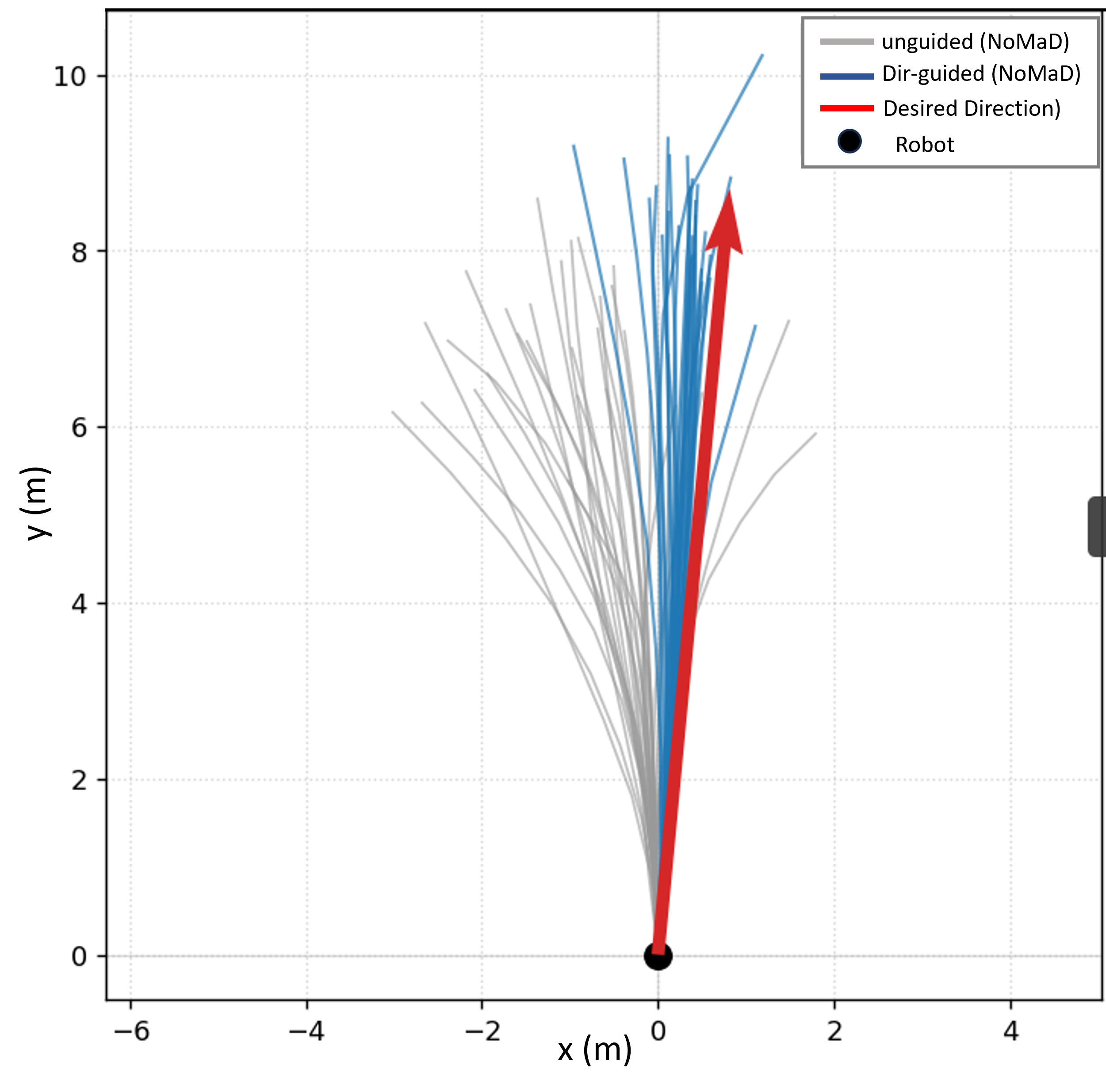}
  \caption{Effect of the direction cost on sampled trajectories, in the robot
  body frame. Without guidance, the goal-masked policy samples a broad,
  multimodal fan of candidates (grey). Applying the gradient of
  Eq.~\eqref{eq:dircost} during the final denoising steps rotates the samples
  toward the desired heading $\hat{\mathbf{d}}$ (red), yielding the
  direction-corrected set (blue).}
  \label{fig:direction_effect}
\end{figure}

In practice, this gradient is obtained by automatic differentiation through
Eqs.~\eqref{eq:denorm}--\eqref{eq:dircost} and injected via the guided update of
Eq.~\eqref{eq:guided_update} only during the final denoising steps, once the
sample is metric enough for a geometric cost to be meaningful. Because the weight
$w$ is small, the guidance \emph{skews} the sampled action distribution toward
$\hat{\mathbf{d}}$ rather than collapsing it onto a single trajectory, thereby
preserving the multimodality that makes exploration useful.
This effect is visible in Fig.~\ref{fig:direction_effect}, where the corrected
samples concentrate around the desired heading while retaining the spread of the
original distribution.
\subsubsection{Traversability Mapping and Best-Direction Selection}
\label{sec:traversability}
 
While direction guidance aligns the exploration mode with the global path toward the goal, it does not account for obstacles that were absent when the environment map was built from the overhead depth camera—and such out-of-distribution obstacles are precisely the cases the default NoMaD policy struggles to avoid. We therefore compute the desired direction $\mathbf{d}^{\mathrm{rel}}$  from an explicit, per-frame \emph{traversability map}, so that the direction handed to the guidance (and to the low-level controller) is at once aligned with the global path and collision-free.
 
\paragraph{Traversability map generation.}
The map generation technique of Sec.~\ref{sec:mapping} is employed here
as well to recover world points from the depth image $D$ produced by the
robot's onboard depth camera. The resulting points are classified by
height $z_r$: those with $\lvert z_r \rvert < \tau_f$ (we use
$\tau_f = 0.08\,$m) are labeled \emph{floor}, while points with
$z_{\min} < z_r < z_{\max}$ (we use $0.1\,$m to $2\,$m) are labeled
\emph{obstacles}; this height gating removes the ground plane, which
would otherwise dominate the projection~\cite{thrun2005}.
Both classes are rasterized into a bird's-eye-view (BEV) grid of
$0.05\,$m resolution centered on the robot. Obstacle cells are inflated
by the robot radius $r_R$ plus a safety margin using a circular
morphological dilation, following standard costmap inflation
practice~\cite{lu2014layered,580977}, so that any cell marked free
is traversable by the robot \emph{center}. The result is a three-state
grid: \emph{verified free} (observed floor outside the inflated margin),
\emph{obstacle} (inflated), and \emph{unknown} (unobserved cells,
including occlusion shadows behind obstacles).
 
\paragraph{Best-direction selection.}
Given the desired direction $\hat{\mathbf{d}}$---computed, as in
Sec.~\ref{sec:direction_cost}, from the relative bearing between the robot's
current pose and the active intermediate waypoint of the global path---we
evaluate a fan of $N$ candidate headings
\begin{equation}
\theta_j \;=\; \angle\hat{\mathbf{d}} + \phi_j,
\qquad \phi_j \in [-90^{\circ},\, 90^{\circ}],
\end{equation}
where $\angle\hat{\mathbf{d}}$ is the desired direction angle and the offsets
$\phi_j$ span $\pm 90^{\circ}$ about it. Each candidate is probed by ray casting
in the BEV traversability grid over a horizon $L$ ($2\,\text{m}$), and
traversability is decided by a \emph{two-tier} rule.

The \emph{hard} rule is never violated: a candidate is rejected if any obstacle
cell is hit within the near range $L_n$ ($1.2\,\text{m}$, the distance traversed
between replanning cycles), either on the ray itself or on two parallel
\emph{side rails} offset by $\pm s$ ($s = 0.15\,\text{m}$). The side rails ensure
the corridor is wide enough for the robot body, preventing the selection of
single-cell slivers that pass a centerline test but leave no lateral clearance.

\begin{figure}[t]
  \centering
  \includegraphics[width=0.8\columnwidth]{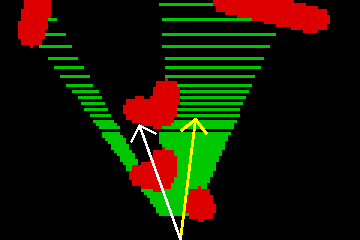}
  \caption{Traversability-based direction selection in the robot body frame.
  The BEV traversability map classifies cells as free (green), occupied (red),
  and unknown (black). The desired direction $\hat{\mathbf{d}}$ (white arrow) is obstructed, so the selector of
  Eq.~\eqref{eq:dirselect} rejects it under the hard rule and returns the
  admissible candidate of smallest angular deviation,
  $\hat{\mathbf{d}}_{\mathrm{trav}}$ (yellow arrow).}
  \label{fig:traversability}
\end{figure}

The \emph{soft} rule then scores every admissible candidate and selects the best:
\begin{equation}
\begin{aligned}
J(\phi_j) &= \lvert \phi_j \rvert + \lambda\, \rho_j, \\
j^\star &= \arg\min_j\, J(\phi_j), \\
\hat{\mathbf{d}}_{\mathrm{trav}} &= \bigl(\cos\theta_{j^\star},\, \sin\theta_{j^\star}\bigr)
\end{aligned}
\label{eq:dirselect}
\end{equation}
where $\hat{\mathbf{d}}_{\mathrm{trav}}$ is the selected \emph{traversable} unit direction—the admissible candidate of smallest angular deviation—obtained from the minimizing heading $\theta_{j^\star}$ and $\rho_j \in [0,1]$ is the fraction of near-range ray cells that are
\emph{unknown} (unobserved) rather than verified free, and $\lambda$ converts
unknown-space exposure into an equivalent angular deviation (we set $\lambda$ so
that $\rho = 1$ is worth $25^{\circ}$). This scoring establishes a strict
preference ordering: at comparable deviation, verified-free directions are
preferred; directions through unknown space (e.g., the occlusion shadow behind an
obstacle) are admissible fallbacks when no verified route exists; and directions
through obstacles are never selected. Because $J$ penalizes angular deviation, the
selector returns the desired direction unchanged whenever it is clear, deviating
only by the smallest angle required for safety as illustrated in Fig.~\ref{fig:traversability}. The selected direction
$\hat{\mathbf{d}}_{\mathrm{trav}} $ is then supplied to the direction-cost guidance of
Sec.~\ref{sec:direction_cost} and to the low-level controller.

\section{Experiments and Results}
\label{sec:experiments}

\subsection{Environment Setup}
\label{sec:env_setup}

All experiments are conducted in a small indoor office environment modelled in
Gazebo, observed by a simulated overhead depth camera mounted on the ceiling. The room measures $9.2 \times 5.2\,\text{m}$ and
is furnished with typical office furniture---tables, chairs, and waste
bin---whose heights span the floor band and obstacle band of Sec.\ref{sec:mapping}. The robot is a TurtleBot~4, whose onboard OAK-D camera provides both RGB and depth. The RGB stream supplies the egocentric observations consumed by NoMaD, while the onboard depth stream is used to compute the per-frame traversability map of Sec.~\ref{sec:guidance}. The overhead depth camera is used exclusively to build the map of Sec.~\ref{sec:mapping} and to compute the global path, and is never available to the policy.

\begin{figure}[t]
  \centering
  \begin{subfigure}[b]{0.49\columnwidth}
    \centering
    \includegraphics[width=\linewidth]{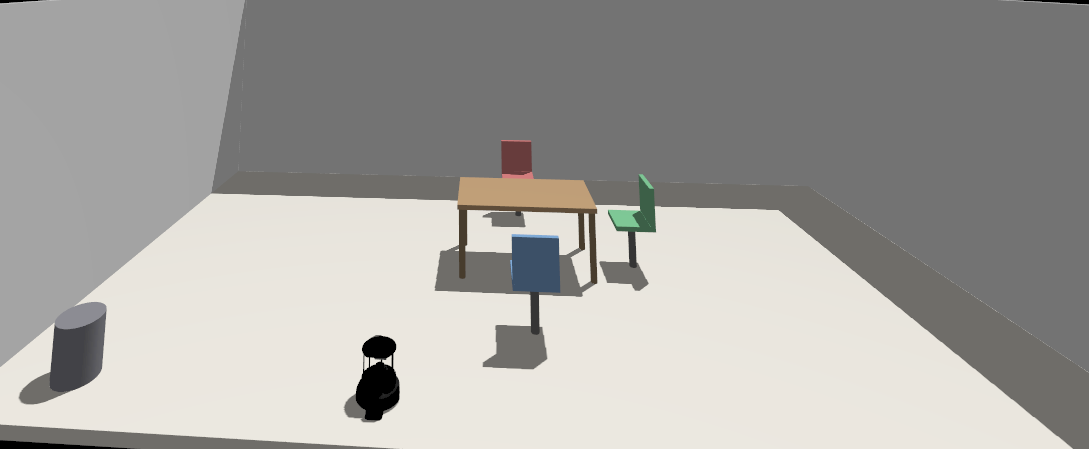}
    \caption{Env-NoObs}
    \label{fig:env_no_obstacles}
  \end{subfigure}\hfill%
  \begin{subfigure}[b]{0.49\columnwidth}
    \centering
    \includegraphics[width=\linewidth]{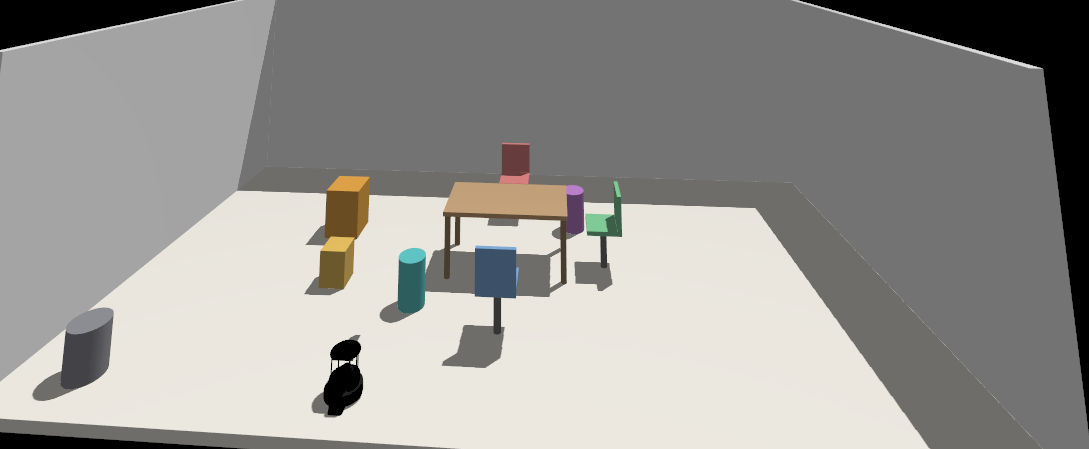}
    \caption{Env-RandObs}
    \label{fig:env_rand_obstacles}
  \end{subfigure}
  \caption{The simulated small office environment in Gazebo.
  (a)~\emph{Env-NoObs} contains only the furniture present when the overhead map
  is built. (b)~\emph{Env-RandObs} additionally contains randomly placed
  obstacles introduced after the map and the global path have been computed.}
  \label{fig:environments}
\end{figure}

We construct two variants of this environment, shown in Fig.~\ref{fig:environments}. The first, \emph{Env-NoObs} (Fig.~\ref{fig:env_no_obstacles}), contains only the furniture present when the overhead map is built; the global path computed on this map is therefore consistent with the scene the robot encounters at run time. The second, \emph{Env-RandObs} (Fig.~\ref{fig:env_rand_obstacles}), additionally contains randomly placed obstacles that are introduced \emph{after} the overhead map has been generated. These obstacles appear in neither the occupancy grid nor the global path, and are precisely the out-of-distribution cases motivating the traversability-based direction selection of Sec.~\ref{sec:guidance}. \emph{Env-NoObs} therefore isolates the effect of direction guidance, while \emph{Env-RandObs} additionally stresses collision avoidance.

We deliberately do not evaluate on established photorealistic simulation
benchmarks such as the Stanford 2D-3D-S \cite{armeni2017} environments, despite their widespread use in visual navigation.
These datasets are reconstructed from egocentric scans of real buildings and
provide no ceiling-mounted vantage point: an overhead depth camera observing the
full workspace from above, which our method requires as its source of global
situational awareness, cannot be instantiated in them without substantially
modifying the scene geometry. 

\subsubsection{Metrics}
To evaluate the state of the art against the proposed ODG-NoMaD, we use four
metrics. \textbf{Success rate} (higher is better) is the fraction of runs in which the robot terminates within
a fixed threshold of the target point $B$ (we use $0.5\,\text{m}$), independently
of whether it collided en route. \textbf{Goal closeness} is the Euclidean
distance between the robot's final position and $B$,
$\lVert \mathbf{p}_{\text{final}} - B \rVert$ (in metres, lower is better). We
report it in addition to success rate because success is binary and coarse: it
cannot distinguish a run that halts just outside the success threshold from one
that wanders off entirely, whereas closeness measures \emph{how far} short a run
terminates and thus captures the graded progress that guidance produces even when
a run is not counted as a success. \textbf{Collision rate} is the fraction of
runs in which the robot contacts an obstacle (lower is better).
\textbf{Success weighted by path length (SPL)}
jointly rewards reaching the goal and doing so efficiently, weighting each success
by the ratio of the shortest feasible path length to the length actually
travelled (higher is better). Each start--goal pair is executed three times per
method, and all metrics are reported as the average over these runs, since the
diffusion policy is stochastic and produces a different rollout on each execution.
\begin{table*}[t]
\centering
\caption{Performance in \emph{Env-NoObs} (no post-mapping obstacles) for the
three start--goal pairs, each averaged over $n=3$ runs. SR: success rate
$\uparrow$; Clo.: goal closeness in metres $\downarrow$; CR: collision rate
$\downarrow$; SPL $\uparrow$. Best value per path--metric in \textbf{bold}.}
\label{tab:env_noobs}
\small
\setlength{\tabcolsep}{4.5pt}
\begin{tabular}{l cccc cccc cccc}
\toprule
& \multicolumn{4}{c}{Path~1} & \multicolumn{4}{c}{Path~2} & \multicolumn{4}{c}{Path~3} \\
\cmidrule(lr){2-5} \cmidrule(lr){6-9} \cmidrule(lr){10-13}
Method & SR & Clo. & CR & SPL & SR & Clo. & CR & SPL & SR & Clo. & CR & SPL \\
\midrule
NoMaD (exploration)       & 0.33 & 0.746 & 1.00 & 0.258 & 0.33 & 0.474 & 1.00 & 0.333 & 0.33 & 1.617 & 1.00 & 0.333 \\
NaviDiffusor (goal guid.) & 0.67 & 0.409 & 1.00 & 0.616 & 0.67 & 0.667 & 0.33 & 0.606 & 0.33 & 1.230 & 0.66 & 0.333 \\
ODG-NoMaD w/o coll.\ avoid.& \textbf{1.00} & 0.127 & \textbf{0.00} & \textbf{1.000} & \textbf{1.00} & 0.187 & \textbf{0.00} & 0.990 & \textbf{1.00} & 0.276 & \textbf{0.00} & \textbf{0.910} \\
ODG-NoMaD (full)          & \textbf{1.00} & \textbf{0.113} & \textbf{0.00} & 0.959 & \textbf{1.00} & \textbf{0.160} & \textbf{0.00} & \textbf{1.000} & \textbf{1.00} & \textbf{0.224} & \textbf{0.00} & 0.887 \\
\bottomrule
\end{tabular}
\end{table*}
 
\begin{table*}[t]
\centering
\caption{Performance in \emph{Env-RandObs}, where randomly placed obstacles are
inserted after the overhead map and the global path have been computed, each
averaged over $n=3$ runs. SR: success rate $\uparrow$; Clo.: goal closeness in
metres $\downarrow$; CR: collision rate $\downarrow$; SPL $\uparrow$. Best value
per path--metric in \textbf{bold}.}
\label{tab:env_randobs}
\small
\setlength{\tabcolsep}{4.5pt}
\begin{tabular}{l cccc cccc cccc}
\toprule
& \multicolumn{4}{c}{Path~1} & \multicolumn{4}{c}{Path~2} & \multicolumn{4}{c}{Path~3} \\
\cmidrule(lr){2-5} \cmidrule(lr){6-9} \cmidrule(lr){10-13}
Method & SR & Clo. & CR & SPL & SR & Clo. & CR & SPL & SR & Clo. & CR & SPL \\
\midrule
NoMaD (exploration)       & 0.00 & 1.054 & 1.00 & 0.000 & 0.00 & 1.229 & 0.66 & 0.000 & 0.33 & 1.110 & 1.00 & 0.333 \\
NaviDiffusor (goal guid.) & 0.67 & 0.382 & 1.00  & 0.629 & 0.67 & 0.699 & 0.33 & 0.591 & 0.00 & 1.246 & 1.00 & 0.000 \\
ODG-NoMaD w/o coll.\ avoid.& \textbf{1.00} & 0.125 & 1.00 & 0.830 & \textbf{1.00} & 0.193 & 1.00 & 0.901 & \textbf{1.00} & \textbf{0.168} & 1.00 & 0.886 \\
ODG-NoMaD (full)          & \textbf{1.00} & \textbf{0.086} & \textbf{0.00} & \textbf{0.994} & \textbf{1.00} & \textbf{0.147} & \textbf{0.00} & \textbf{1.000} & \textbf{1.00} & 0.247 & \textbf{0.00} & \textbf{1.000} \\
\bottomrule
\end{tabular}
\end{table*}

\begin{figure*}[t]
  \centering
  \begin{subfigure}[b]{0.49\linewidth}
    \centering
    \includegraphics[width=\linewidth]{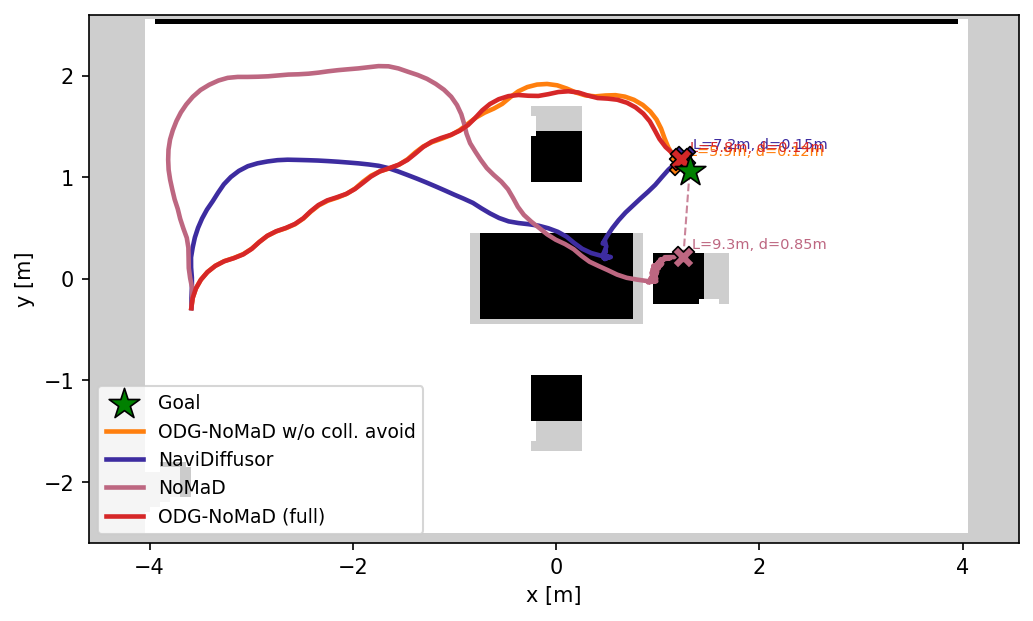}
    \caption{Env-NoObs}
    \label{fig:qual_noobs}
  \end{subfigure}\hfill%
  \begin{subfigure}[b]{0.49\linewidth}
    \centering
    \includegraphics[width=\linewidth]{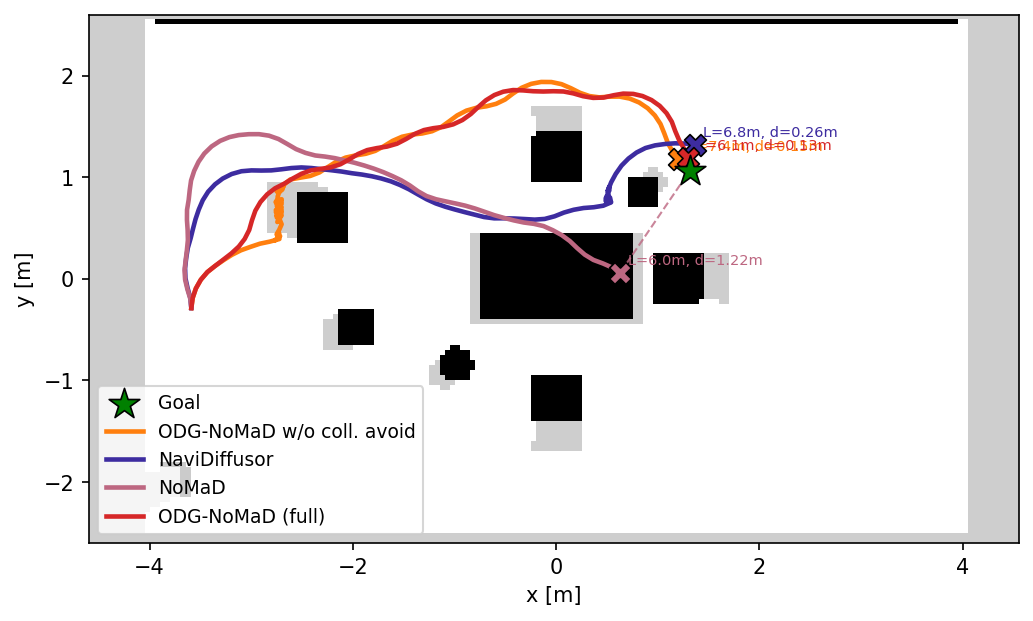}
    \caption{Env-RandObs}
    \label{fig:qual_randobs}
  \end{subfigure}
  \caption{Trajectories executed by the four configurations on Path~1, overlaid on
  the overhead occupancy grid. The green star marks the target $B$; each cross
  marks a run's termination point, annotated with its path length $L$ and goal
  closeness $d$. (a)~In Env-NoObs the ODG-NoMaD trace (red) exactly covers the
  ablated variant (orange), which is therefore not separately visible.
  (b)~In Env-RandObs the two variants separate: the ablated variant tracks the
  path-derived heading into an inserted obstacle, while the full method deviates
  around it.}
  \label{fig:qualitative}
\end{figure*}
\subsubsection{Baselines}
\label{sec:baselines}

We compare four configurations. All of them use the same pretrained NoMaD
checkpoint, without any retraining or fine-tuning; the configurations differ only
in what is applied at inference time.

\textbf{NoMaD (exploration mode).} The unmodified goal-masked diffusion policy
\cite{sridhar2024nomad} operating in its default undirected exploration mode, with no
external guidance. This baseline quantifies the limitation described in
Sec.~\ref{sec:intro}.

\textbf{NaviDiffusor (goal guidance).} Cost-guided diffusion in the spirit of
\cite{navidiffusor}, in which the differentiable goal cost steers the denoising
process toward specified points. We supply the final and intermediate points of
the same global path used by our method, so that both approaches receive
identical global information and differ only in how it is injected.

\textbf{ODG-NoMaD without collision avoidance.} Our direction guidance
(Sec.~\ref{sec:guidance}), in which the desired direction is taken directly from
the active segment of the global path, with the traversability\-/based direction
selection disabled. This isolates the contribution of direction guidance alone.

\textbf{ODG-NoMaD with collision avoidance (full).} Our complete method, in which
the desired direction is obtained from the traversability\-/based selector of
Eq.~\eqref{eq:dirselect}, so that the direction passed to the guidance is both
aligned with the global path and collision-free.

\subsubsection{Experimental Protocol}
\label{sec:protocol}

For each environment we define three start--goal pairs $(A, B)$ spanning the
workspace, chosen to induce global paths of varying geometry: a short direct
route, a route requiring a single turn around furniture, and a longer
multi-segment route traversing most of the room. For each pair, the robot is deployed at the start point $A$
and evaluated under all four configurations of Sec.~\ref{sec:baselines}, using
the same map, path, start, and goal throughout, so that any difference in
performance is attributable solely to the guidance mechanism.

\subsection{Results}
\label{sec:results}

Tables~\ref{tab:env_noobs} and \ref{tab:env_randobs} report success rate (SR),
goal closeness, collision rate (CR), and SPL for the four configurations across
the three start--goal pairs in each environment, averaged over three runs each.

\textbf{Direction guidance recovers goal-directed behaviour.}
Unguided NoMaD rarely reaches the target: it succeeds on only a minority of runs
and terminates well short of the goal, as its exploration wanders without global
direction. Both ODG-NoMaD configurations, in contrast, reach the goal on every
run in both environments, reducing goal closeness by a large margin and raising
SPL accordingly. A directional prior derived from the overhead global path is
thus sufficient to steer the goal-masked policy reliably to the target, without
retraining.

\textbf{Direction guidance outperforms point-goal cost guidance.}
Given the same global information, the NaviDiffusor goal-guidance baseline is
markedly less reliable, succeeding on fewer runs and achieving substantially lower
SPL than ODG-NoMaD. Because its cost gradient pulls each sample toward a fixed
point rather than rotating it along a heading, it is prone to stalling short of
the goal, and on the more demanding routes it fails outright where our method
still succeeds. Steering the policy by direction is therefore more effective than
driving it toward point goals.

\textbf{Traversability-based selection eliminates collisions.}
Collision rate is where the full method separates from its ablation. In
\emph{Env-NoObs}, no obstacle lies beyond those already captured in the overhead map, so
both ODG-NoMaD variants are collision-free and near-identical on every metric,
confirming that the selector returns the path-derived heading unchanged when the
corridor is clear. In \emph{Env-RandObs}, however, the ablated variant---which follows
the path heading without traversability reasoning---collides on obstacles the
planner never saw, whereas the full method remains collision-free. Notably, the
ablated variant still reaches the goal: it succeeds by pushing through obstacles
rather than avoiding them, which is exactly the unsafe behaviour the
traversability selector is designed to remove.

\textbf{Summary.}
Across both environments, direction guidance raises the success rate from a small
fraction to unity and substantially improves SPL over unguided exploration, while
traversability-based selection removes collisions in the presence of unmapped
obstacles---the setting in which its ablation fails. The full ODG-NoMaD is the
only configuration that succeeds on every trial without a single collision, and
all of its guidance is applied at inference, leaving the pretrained NoMaD
checkpoint unchanged.

\textbf{Qualitative behaviour.}
Fig.~\ref{fig:qualitative} overlays the Path~1 trajectories on the overhead
occupancy grid. Unguided NoMaD (pink) drifts through the room and halts in open
space, and NaviDiffusor (blue) is goal-directed but stalls short of the target.
In Env-NoObs (Fig.~\ref{fig:qual_noobs}) the ablated variant (orange) is hidden
beneath the full method (red): with no obstacle beyond those in the overhead map,
the selector of Eq.~\eqref{eq:dirselect} returns the path-derived heading
unchanged, so the two coincide. In Env-RandObs (Fig.~\ref{fig:qual_randobs}) they
separate: the ablated variant follows the path heading into an inserted obstacle
and collides, whereas the full method rejects the blocked headings, bends away by
the smallest admissible angle, and  reach the goal
collision-free and along a shorter route.
 
\section{Conclusion}
\label{sec:conclusion}
 
ODG-NoMaD restores goal-directed behaviour to NoMaD's goal-masked diffusion policy
in unseen environments through a directional prior derived from a global path
planned on an overhead-camera map, together with a traversability-based selector
that keeps the chosen heading clear of obstacles the map never observed---all at
inference time, without retraining the policy. Across both simulated environments,
ODG-NoMaD reaches markedly closer to the target than unguided exploration---cutting
the mean residual distance by roughly $7\times$---and proves more reliable than
NaviDiffusor's point-goal cost guidance. 
Notably, ODG-NoMaD steers the robot around new obstacles placed in the scene after
the overhead map is built, while keeping it on the global path to the goal.

The approach assumes a bounded workspace observed in full by a single calibrated
overhead camera. Larger areas could be covered by fusing several overhead views
into a single map, or by a drone that sweeps the region to build it---leaving the
guidance unchanged, since it depends only on the resulting path. Enriching the map
with semantic labels would further allow targets to be specified as objects rather
than metric coordinates, extending ODG-NoMaD to object-goal navigation.

\bibliographystyle{ACM-Reference-Format}
\bibliography{references}





\end{document}